\documentclass[11pt]{article}

\usepackage{dialogue}

\usepackage{ifxetex}
\ifxetex
    \usepackage{fontspec}
    \setromanfont{Times New Roman}
\else
  \usepackage{cmap}
  \usepackage[T2A,T1]{fontenc}
  \usepackage[utf8]{inputenc}
  \usepackage{times}
  \usepackage{latexsym}
  \usepackage{substitutefont}
  \substitutefont{T2A}{\familydefault}{cmr}
\fi

\usepackage[russian,british]{babel}
\usepackage{url}
\usepackage{pgf}
\usepackage{longtable}
\usepackage{multirow}
\usepackage{makecell}
\usepackage{multicol}
\usepackage{colortbl}
\usepackage{float}
\usepackage{amssymb}

\usepackage{covington} 
\usepackage{booktabs}
\usepackage{hyperref}

\dialogfinalcopy 
\title{Findings of the The RuATD Shared Task 2022 on Artificial Text Detection in Russian}

\author{Tatiana	Shamardina${}^{1}$\thanks{\ \ Equal contribution.}~, Vladislav Mikhailov${}^{2*}$, Daniil Chernianskii${}^{3,4}$,\\
\bf Alena Fenogenova${}^{2}$, Marat Saidov${}^{6}$, Anastasiya Valeeva${}^{7}$, \\
\bf Tatiana	Shavrina${}^{2,3}$, Ivan Smurov${}^{1,7}$, Elena Tutubalina${}^{5,6,8}$, Ekaterina Artemova${}^{6}$ \\
${}^{1}$ABBYY, ${}^{2}$SberDevices, ${}^{3}$AIRI, ${}^{4}$Skolkovo Institute of Science and Technology, ${}^{5}$Sber AI, \\
${}^{6}$HSE University, ${}^{7}$Moscow Institute of Physics and Technology,${}^{8}$Kazan Federal University\\
\small{\textbf{Correspondence}: \href{mailto:tatiana.shamardina@abbyy.com}{tatiana.shamardina@abbyy.com}}
}

\date{}

\begin{document}
\maketitle
\begin{abstract}

We present the shared task on artificial text detection in Russian, which is organized as a part of the Dialogue Evaluation initiative, held in 2022. 
The shared task dataset includes texts from 14 text generators, i.e., one human writer and 13 text generative models fine-tuned for one or more of the following generation tasks: machine translation, paraphrase generation, text summarization, text simplification. We also consider back-translation and zero-shot generation approaches. The human-written texts are collected from publicly available resources across multiple domains.

The shared task consists of two sub-tasks: (i) to determine if a given text is automatically generated or written by a human; (ii) to identify the author of a given text. The first task is framed as a binary classification problem. The second task is a multi-class classification problem. We provide count-based and BERT-based baselines, along with the human evaluation on the first sub-task. A total of 30 and 8 systems have been submitted to the binary and multi-class  sub-tasks, correspondingly. Most teams outperform the baselines by a wide margin. We publicly release our codebase, human evaluation results, and other materials in our \href{https://github.com/dialogue-evaluation/RuATD/blob/main/en_README.md}{GitHub repository}.

  
  \textbf{DOI:} 10.28995/2075-7182-2022-20-XX-XX
\end{abstract}

\selectlanguage{russian}
\begin{center}
  \russiantitle{RuATD-2022: Соревнование по автоматическому распознаванию сгенерированных текстов}
  \medskip

{\bf Татьяна Шамардина}${}^{1*}$, {\bf Владислав Михайлов}${}^{2*}$, {\bf Даниил Чернявский}${}^{3,4}$,\\
{\bf Алена Феногенова}${}^{2}$, {\bf Марат Саидов}${}^{6}$, {\bf Анастасия Валеева}${}^{7}$, \\
{\bf Татьяна Шаврина}${}^{2,3}$, {\bf Иван Смуров}${}^{1,7}$, {\bf Елена Тутубалина}${}^{5,6,8}$, {\bf Екатерина Артемова}${}^{6}$ \\
${}^{1}$ABBYY, ${}^{2}$SberDevices, ${}^{3}$AIRI, ${}^{4}$Сколтех, ${}^{5}$Sber AI, \\ 
${}^{6}$НИУ ВШЭ, ${}^{7}$МФТИ,${}^{8}$КФУ\\
\small{\textbf{Для связи}: \href{mailto:tatiana.shamardina@abbyy.com}{tatiana.shamardina@abbyy.com}}

 \medskip
\end{center}

\begin{abstract}
Данная статья представляет собой отчет организаторов соревнования RuATD-2022, посвященного автоматическому распознаванию сгененированных текстов на материале русского языка. Соревнование RuATD-2022 проходило в рамках кампании Dialogue Evaluation в 2022 году. Набор данных, использованный в соревновании, частично составлен автоматически с использованием моделей генерации текстов. Мы использовали модели, обученные решать различные задачи генерации текстов: машинного перевода, генерация парафразов, автоматического реферирования и упрощения предложений. Мы также рассматриваем популярные постановки задач, такие как обратный перевод и zero-shot генерация. Вторая часть набора данных – тексты, написанные людьми – собрана из открытых источников, относящихся к ряду предметных областей.

Участникам соревнования предлагается решить две задачи: (i) определить, был ли данный текст написан человеком или сгенерирован моделью (бинарная классификация), или (ii) определить автора текста (мультиклассовая классификация). В рамках соревнования мы предоставляем базовые решения в стандартной постановке задачи классификации на основе счетных признаков (TF-IDF) и модели архитектуры BERT. Кроме того, мы проводим оценку решения первой задачи разметчиками на краудсорсинговой платформе (human baseline). В общей сложности, соревнование привлекло внимание 38 решений: 30 для первой постановки задачи и 8 -- для второй. Большая часть участников преодолела уровень базовых решений и уровень разметчиков. Используемая кодовая база, результаты оценки на краудсорсинговой платформе и другие материалы соревнования доступны в публичном  \href{https://github.com/dialogue-evaluation/RuATD/blob/main/en_README.md}{GitHub репозитории соревнования. }

\textbf{Ключевые слова:} распознавание сгенерированных текстов, генерация текстов, соревнование, автоматическое определение автора текста, нейронные сети
\end{abstract}
\selectlanguage{british}

\section{Introduction}
Modern text generative models (TGMs) have demonstrated impressive results in generating texts close to the human level in terms of fluency, coherence, and grammar~\cite{keskar2019ctrl,zellers2019defending,brown2020language,rae2021scaling}. However, the misuse potential of TGMs increases with their capabilities to generate more human-like texts. Malicious users can deploy TGMs for spreading propaganda and fake news~\cite{zellers2019defending,uchendu-etal-2020-authorship,mcguffie2020radicalization}, augmenting fake product reviews~\cite{adelani2020generating}, and facilitating fraud, scams, and other targeted manipulation~\cite{DBLP:journals/corr/abs-2112-04359}. The increasing difficulty for laypeople and users to discriminate machine-generated texts from human-written ones facilitates the spread of such misuse~\cite{karpinska-etal-2021-perils,uchendu-etal-2021-turingbench-benchmark}. This motivates the \textit{artificial text detection} task~\cite{jawahar-etal-2020-automatic}, a fast-growing niche field aimed at mitigating the misuse of TGMs.

The Russian Artificial Text Detection (RuATD) shared task explores the problem of artificial text detection in Russian. Unlike existing datasets for English, our approach includes a range of task-specific TGMs, that is, models fine-tuned for common text generation tasks at the sentence- and document-level. On the one hand, such a setting challenges the participants and crowd-sourced annotators. On the other hand, it also enables many research and development purposes, such as training and benchmarking artificial text detectors, warning users about potentially fake content on social media and news platforms, filtering corpora augmented with TGMs, exploring detectors' robustness w.r.t. TGMs' architecture, size, downstream task, or domain. The shared task dataset consists of publicly available texts across multiple domains and texts generated by various monolingual and cross-lingual TGMs. The setup includes two sub-tasks: (i) to determine if a given text is automatically generated or written by a human (binary classification), and (ii) to identify the author of a given text (multi-class classification).

The main contributions of this paper are the following:
\begin{enumerate}
    \item We propose a diverse automatic text detection corpus in Russian, the first of its kind (\S \ref{subsection:data});

    \item We model two competition sub-tasks (\S \ref{subsection:tasks}) after the traditional concepts of ``Turing test'' and authorship attribution for neural text generation models~\cite{uchendu-etal-2021-turingbench-benchmark}. We establish two count-based and neural-based baseline solutions (\S \ref{subsection: baseline}) and the human evaluation on the binary classification problem (\S \ref{subsection:human benchmark});

    \item We conduct an extensive analysis of the received submissions for both sub-tasks (\S \ref{section:results}) and discuss potential research directions (\S \ref{section:disc});
    
    \item We set up the shared task environment, which remains open for the community submissions to facilitate future research in the area (\S \ref{subsection:kaggle}).
\end{enumerate}

\section{Dataset}
\begin{figure*}[ht!]
    \includegraphics[width=\linewidth]{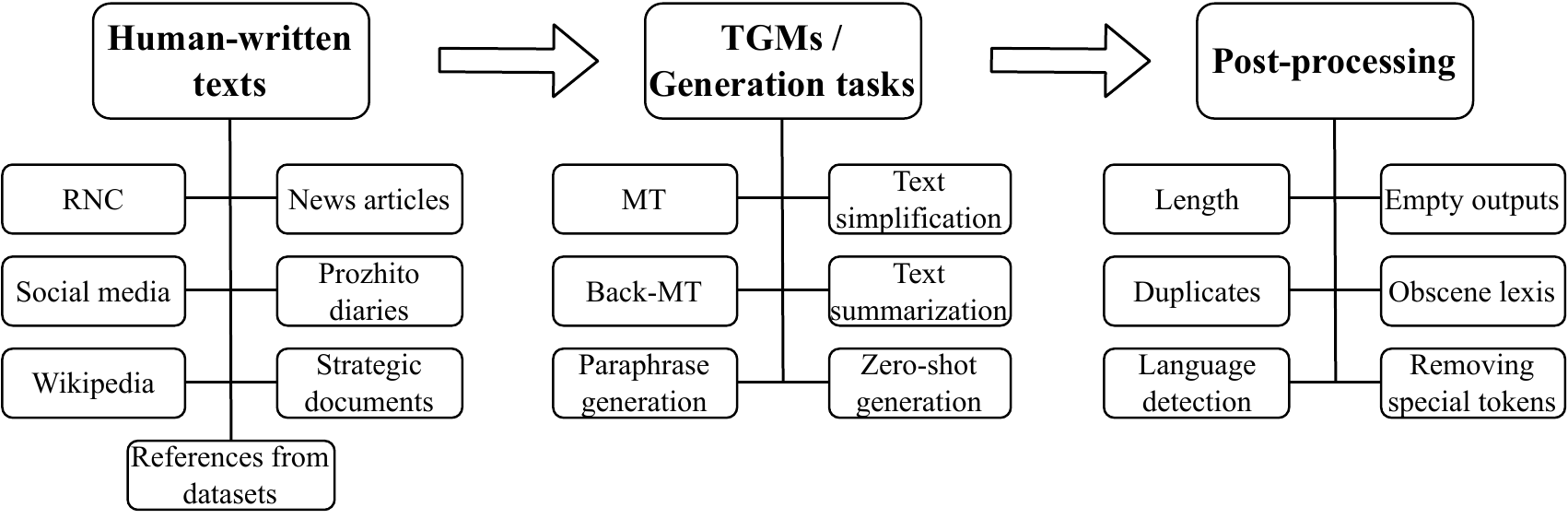}
    \caption{Data collection, text generation, and post-processing procedure.}
    \label{fig:scheme}
\end{figure*}

\subsection{Text Generation}
The corpus includes texts from 14 text generators, i.e., one human writer and 13 monolingual/multilingual TGMs varying in their number of parameters, architecture choices, and pre-training objectives. Each model is fine-tuned for one or more of the following text generation tasks: MT, paraphrase generation, text simplification, and text summarization. We also consider back-translation and zero-shot generation approaches. Figure~\ref{fig:scheme} outlines the dataset creation pipeline. Text generation hyperparameters for each model are presented in \ref{tab:gen-params}. Note that we mostly use the default hyperparameters under the considered libraries.

\begin{table}[ht!]
\centering
\scriptsize
\begin{tabular}{lrr}

\toprule
\textbf{Task} & \textbf{Model} & \textbf{Hyperparameters} \\
\midrule
\multirow{3}{*}{\textbf{Machine Translation}} & OPUS-MT & \multirow{3}{*}{\textsc{Beam search}; num beams=$5$} \\
& M-BART50 \\ & M2M-100  
                                              
\\ \midrule
\multirow{3}{*}{\textbf{Back Translation}} & OPUS-MT & \multirow{3}{*}{\textsc{Beam search}; num beams=$5$} \\
& M-BART50 \\ & M2M-100  \\
\midrule
\multirow{3}{*}{\textbf{Paraphrase Generation}} & mT5-Small, mT5-Large & \multirow{3}{*}{\textsc{top-p sampling}; p=$0.9$} \\

& ruT5-Base-Multitask \\ & ruGPT2-Large, ruGPT3-Large

                                                
\\
\midrule

\multirow{2}{*}{\textbf{Simplification}} & mT5-Large, ruT5-Large & \multirow{2}{*}{\textsc{top-p sampling}; p=$0.9$} \\ 
& ruGPT3-Small, ruGPT3-Medium, ruGPT3-Large 
\\
\midrule
\multirow{2}{*}{\textbf{Summarization}} & M-BART & \multirow{2}{*}{\textsc{Beam search}; num beams=$5$} \\ & ruT5-Base \\

\midrule


\multirow{2}{*}{\textbf{Zero-shot Generation}} & ruGPT3-Small, ruGPT3-Medium & \textsc{top-p sampling}; p=$0.95$; \\ &
ruGPT3-Large & max length=$90$ percentile of length distribution by domain

\\
\bottomrule
\end{tabular}
\caption{A brief description of the text generation hyperparameters and decoding strategies by text generation task.}
\label{tab:gen-params}
\end{table}

\noindent{\bf Human} We collect human-written texts from publicly available resources among six domains (see Section~\ref{subsection:data} for more details). Gold standard references from task-specific datasets are also used as human texts, since they are generally written and/or validated by crowd-source annotators~\cite{artetxe-schwenk-2019-massively,schwenk-etal-2021-wikimatrix,scialom-etal-2020-mlsum,hasan-etal-2021-xl}. The human texts serve as the input to the TGMs.

\noindent{\bf MT \& Back-translation} We use three MT models via the \texttt{EasyNMT} framework\footnote{\href{https://github.com/UKPLab/EasyNMT}{\texttt{github.com/UKPLab/EasyNMT}}}: OPUS-MT~\cite{tiedemann-thottingal-2020-opus}, M-BART50~\cite{tang2020multilingual}, and M2M-100~\cite{fan2020englishcentric}. We use subsets of the Tatoeba~\cite{artetxe-schwenk-2019-massively} and WikiMatrix~\cite{schwenk-etal-2021-wikimatrix} datasets to obtain translations among three language pairs: English-Russian, French-Russian, and Spanish-Russian. In the back-translation setting, the input sentence is translated into one of the target languages, and then back into Russian.

\noindent{\bf Paraphrase Generation} Paraphrases are generated with models available under the \texttt{russian-paraphrasers} library~\cite{fenogenova-2021-russian}: ruGPT2-Large\footnote{\href{https://huggingface.co/sberbank-ai/rugpt2large}{\texttt{hf.co/sberbank-ai/rugpt2large}}}, ruT5-Base-Multitask\footnote{\href{http://huggingface.co/cointegrated/rut5-base-paraphraser}{\texttt{hf.co/cointegrated/rut5-base-multitask}}}, and mT5~\cite{xue-etal-2021-mt5} of Small and Large versions.

\noindent{\bf Text Simplification} We fine-tune ruGPT3-Small\footnote{\href{https://huggingface.co/sberbank-ai/rugpt3small_based_on_gpt2}{\texttt{hf.co/sberbank-ai/rugpt3small}}}, ruGPT3-Medium\footnote{\href{https://huggingface.co/sberbank-ai/rugpt3medium_based_on_gpt2}{\texttt{hf.co/sberbank-ai/rugpt3medium}}}, ruGPT3-Large\footnote{\href{https://huggingface.co/sberbank-ai/rugpt3large_based_on_gpt2}{\texttt{hf.co/sberbank-ai/rugpt3large}}}, mT5-Large, and ruT5-Large\footnote{\href{https://huggingface.co/sberbank-ai/ruT5-large}{\texttt{hf.co/sberbank-ai/rugt5-large}}} for text simplification on a filtered version of the RuSimpleSentEval-2022 dataset~\cite{sakhovskiy2021rusimplesenteval,fenogenovatext}. Fine-tuning of each model is run for $4$ epochs with the batch size of $4$, learning rate of $10^{-5}$, and weight decay of $10^{-2}$.

\noindent{\bf Text Summarization} We use two abstractive summarization models fine-tuned on the Gazeta dataset~\cite{Gusev2020gazeta}: ruT5-base\footnote{\href{https://huggingface.co/IlyaGusev/rut5_base_sum_gazeta}{\texttt{hf.co/IlyaGusev/rut5-base-sum-gazeta}}} and M-BART\footnote{\href{https://huggingface.co/IlyaGusev/mbart_ru_sum_gazeta}{\texttt{hf.co/IlyaGusev/mbart-ru-sum-gazeta}}}.  

\noindent{\bf Zero-shot Generation} We generate texts in a zero-shot manner by prompting the model and specifying the maximum number of generated tokens. The models include ruGPT3-Small, ruGPT3-Medium, ruGPT3-Large.

\subsection{Data}
\label{subsection:data}
Pre-training corpora of TGMs can cover multiple versatile domains~\cite{liu2020survey}, which prompt their abilities to generate texts with specific lexical, syntactic, discourse and stylistic properties. Despite this, the ATD task is generally explored w.r.t. only one particular domain, e.g., product reviews ~\cite{adelani2020generating}, social media posts~\cite{fagni2021tweepfake}, or news~\cite{uchendu-etal-2021-turingbench-benchmark}. Such setting limits the scope of evaluation of artificial text detectors. A few studies show that performance of modern detectors can vary drastically across domains~\cite{bakhtin2019real,kushnareva-etal-2021-artificial}, which stimulates the development of more generalizable and robust methods~\cite{jawahar-etal-2020-automatic}.

This paper aims at providing a diverse shared task data, taking into account the current limitations in the niche ATD field, and the diversity of TGMs widely used in the industry and NLP research for Russian. To this end, we consider domains which represent normative Russian, as well as general domain texts, social media posts, texts of different historical periods, bureaucratic texts with complex discourse structure and embedded named entities, and other domains included in the task-specific datasets, such as subtitles and web-texts. Recall that aside from linguistic and stylometric properties, texts differ in their length (e.g., sentence-level vs. document-level), and specifics attributable to the downstream tasks. We now list domains of texts that are fed into the previously described TGMs.

\noindent{\bf Russian National Corpus} We use the diachronic sub-corpora of the Russian National Corpus\footnote{\href{https://ruscorpora.ru/old/en/index.html}{\texttt{ruscorpora.ru}}} (RNC), which covers three historical periods of the society and the Modern Russian language (``pre-Soviet'', ``Soviet'', and ``post-Soviet'').

\noindent{\bf Social Media} We parse texts from multiple social media platforms that are marked with certain hashtags, such as dates, months, seasons, holidays, the names of large cities in Russia, etc. These texts are typically short, written in informal style and may contain emojis and obscene lexis.

\noindent{\bf Wikipedia} We select the top-100 most viewed Russian Wikipedia pages spanning the period of 2016-2021 according to the PageViews\footnote{\url{https://pageviews.wmcloud.org/}} statistics. 

\noindent{\bf News Articles} The news segment covers different news sources in the Taiga corpus~\cite{shavrina2017methodology} and the \texttt{corus} library\footnote{\href{https://github.com/natasha/corus}{\texttt{github.com/natasha/corus}}}, including but not limited to Lenta, KP, Interfax, Izvestia, Gazeta. We additionally parse more recent news articles to prevent potential data leakage and cheating.

\noindent{\bf Prozhito Diaries} Prozhito is a corpus of digitilized personal diaries, written during the 20th century~\cite{melnichenko2017prozhito}.

\noindent{\bf Strategic Documents} are produced by the Ministry of Economic Development of the Russian Federation. The documents are written in bureaucratic style, rich in embedded entities, and have complex syntactic and discourse structure. This dataset has been previously used in the RuREBus shared task \cite{rurebus}.

\subsection{Post-processing} Each generated text undergoes a post-processing procedure based on a combination of language processing tools and heuristics. First, we discard duplicates, copied inputs, empty outputs, and remove special tokens from the generated texts (e.g., \texttt{<s>}, \texttt{</s>}, \texttt{<pad>}, etc.). Next, we empirically define length intervals for each generation task based on a manual analysis of length distributions in \texttt{razdel}\footnote{\href{https://github.com/natasha/razdel}{\texttt{github.com/natasha/razdel}}} tokens. The texts are filtered by the following token ranges: 5-to-25 (\textbf{MT, Back-translation, Paraphrase Generation}), 10-to-30 (\textbf{Text Simplification}), 15-to-60 (\textbf{Text Summarization}), and 85-to-400 (\textbf{Zero-shot Generation}). We additionally discard the social media texts containing obscene lexis according to the corpus of Russian obscene words\footnote{\href{https://github.com/odaykhovskaya/obscene_words_ru}{\texttt{github.com/odaykhovskaya/obscene-words}}}, and keep the MT/Back-translation texts classified as Russian with the confidence of more than $0.9$ (\texttt{langdetect}\footnote{\href{https://github.com/fedelopez77/langdetect}{\texttt{github.com/fedelopez77/langdetect}}}).


\section{Dataset Statistics}
This section describes various count-based statistics of our dataset for human-written and machine-generated texts.

\noindent{\bf General Statistics} \autoref{tab:tgms} shows general dataset statistics w.r.t. text generation task, text generator, and domain. On average, there are $37.9$ tokens in each text, with variations depending on the task. We estimate the frequency of each text according to the Russian National Corpus (RNC)\footnote{\href{http://ruscorpora.ru/new/en/}{\texttt{ruscorpora.ru/new/en}}}. It is computed as the number of frequently used tokens (i.e., the number of instances per million, that is, IPM in RNC is higher than $1$) divided by the number of tokens in a sentence. The average IPM is $0.86$ for the human-written texts and $0.87$ for the machine-generated ones.

\begin{table*}[t!] 
    \centering
    \scriptsize
    \begin{tabular}{cccccc}

    \toprule
    \textbf{Task} & \textbf{Text Generator} & \textbf{Domain} & \textbf{Num. samples} & \textbf{Num. tokens} & \textbf{IPM} \\
    \midrule

    
    \textbf{Machine Translation} &
        \begin{tabular}{@{}c@{}c@{}c@{}}
            Human \\ OPUS-MT \\ M-BART50 \\ M2M-100
        \end{tabular} &
        
        \begin{tabular}{@{}c@{}}
            Tatoeba \\ WikiMatrix  \\
        \end{tabular} &
        
        \begin{tabular}{@{}c }
            35860
        \end{tabular} &
        
        \begin{tabular}{@{}c }
            11.5
        \end{tabular} &
        
        \begin{tabular}{@{}c }
            0.89
        \end{tabular}
        
    \\\midrule
    
    \textbf{Back Translation} &
        \begin{tabular}{@{}c@{}c@{}c@{}}
            Human \\ OPUS-MT \\ M-BART50 \\ M2M-100 
        \end{tabular} &
        \begin{tabular}{@{}c@{}c@{}c@{}c@{}c@{}c@{}}
            Strategic documents \\ News \\ Prozhito \\ RNC \\ Wikipedia \\ Tatoeba \\ WikiMatrix 
        \end{tabular} &
        
        \begin{tabular}{@{}c }
            35588
        \end{tabular} &
        
        \begin{tabular}{@{}c }
            12.9
        \end{tabular} &
        \begin{tabular}{@{}c }
            0.88
        \end{tabular}
    
    \\\midrule
    
    \textbf{Paraphrase Generation} &
        \begin{tabular}{@{}c@{}c@{}c@{}c@{}c@{}}
            Human \\ mT5-Small \\ mT5-Large \\ ruT5-Base-Multitask \\ ruGPT2-Large \\
            ruGPT3-Large
        \end{tabular} &
        \begin{tabular}{@{}c@{}c@{}c@{}c@{}c@{}}
            Strategic documents \\ News \\ Prozhito \\ RNC \\ Wikipedia \\ Social media 
        \end{tabular} &
        
        \begin{tabular}{@{}c }
            44298
        \end{tabular} &
        
        \begin{tabular}{@{}c }
            13.0
        \end{tabular} &
        
        \begin{tabular}{@{}c }
            0.85
        \end{tabular}
    
    \\\midrule
    
    \textbf{Simplification} &
        \begin{tabular}{@{}c@{}c@{}c@{}c@{}c@{}}
            Human \\ mT5-Large \\ ruT5-Large \\ ruGPT3-Small \\ ruGPT3-Medium \\ ruGPT3-Large
        \end{tabular} &
        \begin{tabular}{@{}c@{}c@{}c@{}c@{}c@{}}
        Strategic documents \\ News \\ Prozhito \\ RNC \\ Wikipedia \\ Social media 
        \end{tabular} &
        
        \begin{tabular}{@{}c }
            44700
        \end{tabular} &
        
        \begin{tabular}{@{}c }
            18.3
        \end{tabular} &
        
        \begin{tabular}{@{}c }
            0.86
        \end{tabular}
    
    \\\midrule
    
    \textbf{Summarization} &
        \begin{tabular}{@{}c@{}c@{}}
            Human \\ M-BART \\ ruT5-Base
        \end{tabular} &
        \begin{tabular}{@{}c@{}c@{}c@{}c@{}}
        Strategic documents \\ News \\ Prozhito \\ RNC \\ Wikipedia 
        \end{tabular} &
        
        \begin{tabular}{@{}c }
            17164
        \end{tabular} &
        
        \begin{tabular}{@{}c }
            33.5
        \end{tabular} &
        \begin{tabular}{@{}c }
            0.86
        \end{tabular}
    
    \\\midrule
    
    \textbf{Zero-shot Generation} &
        \begin{tabular}{@{}c@{}c@{}c@{}}
            Human \\ ruGPT3-Small \\ ruGPT3-Medium \\ ruGPT3-Large
        \end{tabular} &
        \begin{tabular}{@{}c@{}c@{}c@{}c@{}}
        Strategic documents \\ News \\ Prozhito \\ RNC \\ Wikipedia 
        \end{tabular} &
        
        \begin{tabular}{@{}c }
            37499
        \end{tabular} &
        
        \begin{tabular}{@{}c }
            141.5
        \end{tabular} &
        
        \begin{tabular}{@{}c }
            0.85
        \end{tabular}
    
    \\\bottomrule
\end{tabular}
\caption{Text generators, domains and the final number of samples per task. The number of human-written texts is same as machine-generated texts.}

\label{tab:tgms}


\end{table*}

\noindent{\bf Diversity Metrics} We estimate the diversity of the texts in terms of their k-gram statistics and lexical richness. We calculate two diversity metrics upon $k$-gram statistics: Dist-$k$~\cite{li-etal-2016-diversity} and Ent-$k$~\cite{zhang2018generating}. Dist-$k$ is the total number of k-grams divided by the number of tokens in the text set. Ent-$k$ is an entropy metric that weights each k-gram so infrequent k-grams are penalized and contribute less to diversity. We compute the diversity scores for texts grouped by label and report them for $k \in \{1, 2, 4\}$ in \autoref{tab:dist-ent}.

 \begin{table*}
\centering
\scriptsize

\begin{tabular}{ll|c|c}
\toprule
  &   & \textbf{H} & \textbf{M} \\
 & \textbf{k} &           &            \\
\midrule
     & \textit{1} &  0.35 &   {\bf 0.40} \\
\textbf{Dist-k}   & \textit{2} &  0.75 &   {\bf 0.76} \\
     & \textit{4} &  0.74 &   {\bf 0.77} \\
\midrule
     & \textit{1} &  {\bf 7.86} &   7.30 \\
\textbf{Ent-k}    & \textit{2} & {\bf 9.99} &   9.03 \\
     & \textit{4} &  {\bf 10.16} &  9.19 \\
\bottomrule
\end{tabular}
\caption{Dist-\textit{k} and Ent-\textit{k} diversity measures by the target level. \textbf{H}=Human-written texts; \textbf{M}=Machine-generated texts.}
\label{tab:dist-ent}
\end{table*}

\noindent{To} measure the lexical diversity\footnote{\href{https://github.com/LSYS/LexicalRichness}{\texttt{Lexical richness}}} of the texts in our dataset, we calculate four types of metrics: word count, terms count, type-token ratio (TTR), and corrected type-token ratio (CTTR). Type-token ratio is computed as $t/w$ and corrected type-token ratio is computed as $t/\sqrt{2 * w}$, where $t$ is the number of unique terms/vocabulary, and $w$ is the total number of words.

\begin{table*}[t!]
\centering
\scriptsize
\begin{tabular}{l|*{14}{c}}
\toprule
 & \multicolumn{2}{c}{{\bf Back-MT}} & \multicolumn{2}{c}{{\bf MT}} & \multicolumn{2}{c}{{\bf Zero-shot Gen.}} & \multicolumn{2}{c}{{\bf Paraphrase Gen.}} & \multicolumn{2}{c}{{\bf Simplification}} & \multicolumn{2}{c}{{\bf Summarization}} & \multicolumn{2}{c}{{\bf Overall}} \\
\midrule
 & H & M & H & M & H & M & H & M & H & M & H & M & H & M\\
\midrule
{\bf Words} & 10.04 & {\bf 10.66} & {\bf 9.65} & 8.84 & 106.05 & {\bf 112.48} & 10.21 & {\bf 11.64} & 14.48 & {\bf 14.70} & 24.46 & {\bf 30.34} & 28.82 & {\bf 30.72} \\
{\bf Terms} & 9.70 & {\bf 10.07} & {\bf 9.33} & 8.47 & 73.52 & {\bf 95.40} & 9.87 & {\bf 11.33} & {\bf 13.72} & 13.55 & 22.32 & {\bf 26.16} & 22.64 & {\bf 26.95} \\
{\bf TTR} & 0.95 & 0.95 & {\bf 0.98} & 0.97 & 0.70 & {\bf 0.86} & 0.96 & {\bf 0.97} & {\bf 0.95} & 0.93 & {\bf 0.93} & 0.87 & 0.91 & {\bf 0.93} \\
{\bf CTTR} & 2.09 & {\bf 2.13} & {\bf 2.08} & 1.97 & 4.91 & {\bf 6.20} & 2.13 & {\bf 2.31} & {\bf 2.52} & 2.47 & 3.14 & {\bf 3.33} & 2.76 & {\bf 3.01} \\
\bottomrule
\end{tabular}
\caption{Lexical richness metrics per text generation task.}
\label{tab:lex-rich}
\end{table*}

We can see that the ratio of the diversity measures between the natural and artificial texts depends on the task, which is explained by the very task formulation. At the same time, artificial texts may include non-existent words, degenerated textual segments, or rare words, which can be attributed to more significant lexical richness metrics overall.

\section{Setup}

\subsection{Tasks} \label{subsection:tasks}

The RuATD Shared task features two sub-tasks: 
\begin{enumerate}
    \item[I.] Determine if a given text is automatically generated or written by a human. This sub-task is framed as a binary classification problem with two labels: \textbf{H} (human) and \textbf{M} (machine).
    \item[II.] Identify the author of a given text. This sub-task is modeled after the traditional problem of authorship attribution~\cite{coyotl2006authorship}, particularly in the context of neural models~\cite{uchendu-etal-2020-authorship}. It is a multi-class classification problem with 14 target classes -- a human writer and 13 TGMs.
\end{enumerate}

\noindent{\bf Evaluation} Each sub-task uses the accuracy score, a standard metric for classification with balanced classes, as the official evaluation metric.

\subsection{Dataset Splits}
We split the dataset into four sets in the 60/10/15/15 proportion ratio: train (130k), development (21k), public test (32k), and private test (32k). Each set is balanced by the number of target classes, text generator, text generation task, and domain\footnote{The number of human-written texts is equal to the number of machine-generated texts for each domain and text generation task.}. These sets are used for both sub-tasks, with only the target classes changed, i.e., the \textbf{M} label is broken into 13 TGMs' names in the multi-class sub-task. 

\subsection{Kaggle Setup} \label{subsection:kaggle}

We use the Kaggle competition platform to run the shared task. The sub-tasks are set as separate competitions and leaderboards: 
\begin{enumerate}
    \item[I.] The binary sub-task \href{https://www.kaggle.com/competitions/ruatd-2022-bi/host/settings}{is hosted under this link};
    \item[II.] The multi-class sub-task \href{https://www.kaggle.com/competitions/ruatd-2022-multi-task}{is hosted under this link}.
\end{enumerate}

The participants are allowed to take part solely or in teams in both sub-tasks. The shared task comprises two stages: \textbf{public} and \textbf{private} testing. The first stage provides access to the public test set and leaderboard, allowing the participants to develop and improve their submissions during the competition. The second stage defines the final leaderboard ranking on the private test set, scoring up to three submissions selected by the participants. Otherwise, the Kaggle platform automatically selects the three best submissions based on the participants' public test scores. 
Participants are allowed to use any additional materials and pre-trained models, except for direct markup of the test set and search on the Internet.

\subsection{Baseline}
\label{subsection: baseline}
We provide the participants with two open-source baseline solutions: count-based (TF-IDF baseline) and BERT-based (BERT baseline)~\cite{devlin-etal-2019-bert}. TF-IDF baseline is based on TF-IDF features coupled with the SVD dimensionality reduction and a Logistic Regression classifier. The TF-IDF has 50k features, further reduced to 5000 by SVD. The BERT baseline follows the default fine-tuning and evaluation procedure for the classification task under the HuggingFace transformers framework~\cite{wolf-etal-2020-transformers}.

\subsection{Peer Review}
Each participant is asked to publicly release their solutions and peer review other participants' submissions. This step allows for a fair evaluation, eliminating the risks of potential cheating, such as solving the sub-tasks via a web search or other heuristics. After analyzing the assigned submission, the peer-reviewer should answer two questions in the Google form and provide comments, if any:

\begin{itemize}
    \item Does the submission use a web search?
    \item Does the submission violate any other rules\footnote{The shared task rules are provided \href{https://github.com/dialogue-evaluation/RuATD/blob/main/en_README.md}{in the GitHub repository}.} of the shared task?
\end{itemize}

\subsection{Human Baseline}
\label{subsection:human benchmark}

\begin{figure*}[t!]
    \includegraphics[width=\linewidth]{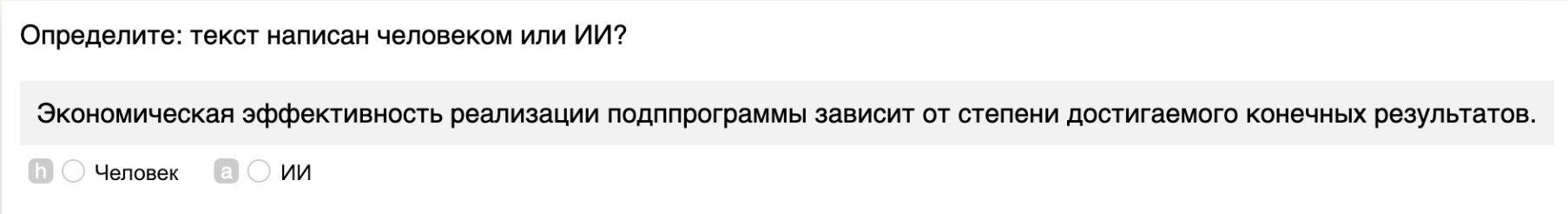}
    \caption{An example of the Toloka interface for the human evaluation setup.}
    \label{fig:tolokasetup}
\end{figure*}

We conduct a human evaluation on the binary classification problem using stratified sub-samples from the public and private test sets. Each subset of 2.5k samples is balanced by the number of target classes, text generator, text generation task, and domain. We report the human baseline results during both public and private testing stages. The evaluation is run via the Toloka platform. The annotation setup follows the conventional crowd-sourcing guidelines for the ATD task and accounts for methodological limitations discussed in ~\cite{ippolito2020automatic,clark-etal-2021-thats,karpinska-etal-2021-perils}. We provide a full annotation instruction in Figure~\ref{fig:toloka_instruction} (see Appendix~\ref{section:appendix}), and an example of the Toloka interface in Figure~\ref{fig:tolokasetup}.

We grant access to the human evaluation project to only top-$70$\% annotators according to the in-house Toloka rating system. Each annotator must first finish the training task by completing at least $80$\% of samples correct to get onto the main annotation task. We use the dynamic overlap of $3$-to-$5$ annotators per sample. We discard votes from those annotators whose quality rate on the control tasks is less than $50$\%. We also filter out votes with the response time of less than $15$ seconds per annotation task page (5 samples). The resulting vote is aggregated as the majority vote label.

\section{Results and Analysis}
\label{section:results}


We report the official shared task results of the private testing stage in \autoref{tab:leaderboards}. Only peer-reviewed submissions (marked with \checkmark in the Table) are considered official.

 \begin{table*}[t!]



 \centering
 \scriptsize
\begin{tabular}{lr|lr}
\toprule
\multicolumn{2}{c|}{\textbf{Binary sub-task}}  & \multicolumn{2}{c}{\textbf{Multi-class sub-task}} \\
\textbf{Team} &    \textbf{Accuracy} & \textbf{Team} & \textbf{Accuracy} \\
\midrule

MSU \checkmark &  0.82995 & Posokhov Pavel \checkmark  &  0.65035\\
Igor &  0.82725 & Yixuan Weng \checkmark  &  0.64731 \\
Orzhan \checkmark  &  0.82629 & Orzhan \checkmark  &  0.64573 \\
mariananieva \checkmark &  0.82427 & MSU \checkmark &  0.62856\\ 
Ivan Zakharov &  0.82294 & \cellcolor[gray]{0.8}BERT baseline  &   \cellcolor[gray]{0.8} 0.59813\\ 
Yixuan Weng \checkmark  &  0.81767 & Nikita Selin &  0.58967\\
ilya koziev &  0.81699 & Victor Krasilnikov &  0.55012\\
miso soup \checkmark  &  0.81178 & Petr Grigoriev \checkmark  &  0.45814\\
Eduard Belov &  0.80862 &\cellcolor[gray]{0.8}TF-IDF baseline &  \cellcolor[gray]{0.8}0.44280\\
Posokhov Pavel \checkmark &  0.80630 & Anastasiya Shabaeva &  0.05411\\ Kirill Apanasovich &  0.80308 \\
Tumanov Alexander &  0.79778 \\ 
\cellcolor[gray]{0.8}BERT baseline &  \cellcolor[gray]{0.8} 0.79666 \\ 
     Elizaveta Nosova &  0.79595 \\
              mipatov &  0.78591 \\
               akstar &  0.78442 \\
         Nikita Selin &  0.78228 \\
        David Avagyan &  0.77869 \\
      Mikhail Yumanov &  0.77181 \\
    Gregory Kuznetsov &  0.75237 \\
   Anastasiya Shabaeva &  0.75178 \\
           Shershunya &  0.74534 \\
 Ekaterina Kostrykina &  0.74326 \\
   Victor Krasilnikov &  0.74091 \\
                Alena &  0.73589 \\
 Alexander Tesemnikov &  0.73204 \\
           Lera Lelik &  0.72727 \\
    Dmitriy Vahrushev &  0.71559 \\
    \cellcolor[gray]{0.8}Human baseline    &   \cellcolor[gray]{0.8} 0.66666 \\
      Molostvov Pavel &  0.68543 \\
           Mental Sky &  0.65326 \\
       Petr Grigoriev \checkmark  &  0.64232 \\ 
      \cellcolor[gray]{0.8}TF-IDF baseline  &  \cellcolor[gray]{0.8} 0.64223 \\
\bottomrule
\end{tabular}
\caption{The official shared task results sorted in the descending order. Left: the binary leaderboard; Right: the multi-class leaderboard. Baseline submissions are colored in grey. \checkmark stands for peer-reviewed submissions.}
\label{tab:leaderboards}


\end{table*}

As one can notice from Table, top-4 systems in the binary classification task have been peer-reviewed. In the multi-class setting, all four top-4 systems have been also peer-reviewed.

The results demonstrate that state-of-the-art classification models can be relatively successful in distinguishing human-written texts from machine-generated ones for the Russian language and determining the exact model used for generation for the latter class. However, one can quickly notice a rather stark contrast between the best scores obtained on the RuATD test set in binary setup (0.830 accuracy for \textit{MSU}, the top-system in binary classification task) and scores obtained for a similar setup in English (0.970 accuracy; see ~\cite{uchendu-etal-2020-authorship} for reference). We attribute this contrast not to the difference in the languages but mainly to the nature of texts: in the English setup, an average text length is 432 words (compared to 31 in RuATD). This claim can be validated by splitting evaluation scores of the best binary RuATD models by length: on the texts longer than 23 words (about a quarter of all RuATD texts), top models can score over 0.95 accuracy.

Unsurprisingly all models that can outperform our BERT baseline used fine-tuned language models (LMs) from the BERT family. Specific models that can achieve the best scores on the test set are mDeBERTa~\cite{he2021debertav3}, and Russian-language implementations of RoBERTa~\cite{liu2019roberta}. Top models experiment with learning-rate scheduling as well as other training techniques (e.g., adversarial training with fast-gradient method ~\cite{dong2018boosting}, or child-tuning training ~\cite{xu2021raise}).

Using additional features (e.g., lexical richness, perplexity, number of characters, number of sentences, TF-IDF of POS tags, punctuation, tonality, reading ease) provided only limited benefit. While there are competitive solutions with such features (e.g., \textit{mariananieva}, 4th-placed solution in the binary setup), none of the three best models in either task used any additional features.

Ensembling models proved to be beneficial, although competitive results could be achieved using single models. For example, \textit{Posokhov Pavel}, the best model in the multi-class setup task, does not use the ensembling of any kind, nor does \textit{orzhan}, the third-placed model in both tasks. 

\noindent{\textbf{Human Baseline}} The overall accuracy of the human evaluation is $0.66$, which scores below the BERT baseline. The low results are consistent with recent studies~\cite{karpinska-etal-2021-perils,uchendu-etal-2021-turingbench-benchmark}, which underpin the difficulty of the task for crowd-sourcing annotators. These works advise hiring experts trained to evaluate written texts or conduct multiple crowd-sourcing evaluation setups with extensive training phases. We leave the human evaluation experiments for future work. 

\section{Discussion}
\label{section:disc}
\noindent{\textbf{On indistinguishable examples}} The reasons for the errors of various systems on the RuATD corpus are of separate research interest. A short meaningful sentence of frequency n-grams may often occur in a web-corpus and be easily reproduced by a simple statistical LM. Thus, the very definition of a specific automatic text can be a challenging task for an attentive annotator and even for an engineer directly involved in developing TGMs. This can be illustrated, for example, by the case of Ilya Sutskever from the GPT-3 project, who tweeted spring of 2022, that \textit{large neural networks may be ``slightly conscious.''}\footnote{\href{https://towardsdatascience.com/openais-chief-scientist-claimed-ai-may-be-conscious-and-kicked-off-a-furious-debate-7338b95194e}{https://towardsdatascience.com}}. The methodological problem of obtaining some significant phrases or texts randomly using LMs, however, is raised much earlier than the onset of ``indistinguishability by the engineers themselves'': critical works on the Turing test~\cite{turing1950computing} offer various variations of tests that level this problem. For example, \cite{bringsjord1996inverted} explicitly note that a state machine that generates random sentences could be randomly considered meaningful by a judge in a good mood.
In general, various methodological variations offer 1) interactive work with models/people, checking the maintenance of the context~\cite{kugel1990time} and even the consistency of the author's ``cognitive profile''~\cite{watt1996naive}. These areas can be considered topics for future work for the following shared tasks.

\noindent{\textbf{Ethical considerations}} Setting the task of detecting non-human texts is timely due to the rapid development of LMs. The very issue of detecting non-human texts affects the fundamental right of the user to understand when they interact with a subjectless technological solution and when - with a person. Problems of this kind are actively discussed in reviews of recent years. In particular, \cite{https://doi.org/10.48550/arxiv.2108.07258} define the scope of problems as: \begin{enumerate}
   \item foundation model misuse, including both purposeful generated text misuse and the unconditional reliance on automatic text classification results that can be false negative;
   \item development of legal grounds to mitigate generative model misuse and detection model misuse;  
\item widespread deployment of automatic text detection systems: the presented models can lead to an "arms race" between malicious content generators and detectors.
\end{enumerate}

Although the improvement of language modeling is undoubtedly a fundamental task of machine learning, we are of the position that a thorough study of models that classify automatic texts is necessary. As practice shows, the percentage of their errors in the Russian language is non-zero.

\section{Related Work}
Many research efforts are related to natural language generation (NLG) models. These works can be characterized into two broad categories - (i) training LMs on large-scale data and (ii) learning to distinguish between machine-authored and human-written content. \cite{jawahar-etal-2020-automatic} provides a good survey on the automatic detection of machine-generated text for English. 

Prior work has focused on training classifiers on samples from a model \cite{brown2020language} and directly using a model distribution \cite{gehrmann2019gltr}. \cite{gehrmann2019gltr} propose a visual and statistical tool named GLTR for the detection of generation artifacts across different sampling schemes. \cite{ippolito2020automatic} compare human raters and automatic classifiers depending on the decoding strategy. They observe that classifiers can detect statistical artifacts of generated sequences while humans quickly notice semantic errors. Classifier accuracy ranges between 70\% and 90\% depending on the decoding strategy for short texts (64 tokens). \cite{dugan2020roft} propose a RoFT (Real or Fake Text) tool to detect the boundary between a human-written text passage and machine-generated sentences showing NLG models are capable of fooling humans by one or two sentences. A recent study of \cite{galle2021unsupervised} focuses on the unsupervised detection of machine-generated documents leveraging repeated higher-order n-grams. They show that specific well-formed phrases over-appear in machine-generated texts as compared to human ones. \cite{mccoy2021much} propose a suite of analyses called RAVEN for assessing the novelty of generated text, focusing on sequential structure (n-grams) and syntactic structure. Experiments show that random sampling result in generated text with a more significant number of novel n-grams.

Recent studies \cite{carlini2022quantifying,lee2022language} have raised a concern about model memorization due to data privacy leakage. \cite{carlini2022quantifying} confirm that memorization scales with model size and current LMs do accurately model the distribution of their training data. \cite{lee2022language} investigate memorization and plagiarism when generating artificial texts. They observe that fine-tuned LMs demonstrate different patterns of plagiarism based on characteristics of auxiliary data. \cite{schuster2020limitations} propose two benchmarks demonstrating the stylistic similarity between malicious and legitimate uses of LMs.

\cite{liyanage2022benchmark} propose a benchmark for detecting
automatically generated research content that consists of a synthetic dataset and a partial text substitution dataset. The latter is created by replacing several sentences of abstracts with sentences generated by an NLG model. \cite{stiff2021detecting} adopt a wide variety of datasets of news articles, product reviews, forum posts, and tweets and investigated several classifiers to predict whether a text has been automatically generated. Their experiments show that classifiers perform reasonably accurately in the news domain, while the same task is more challenging for shorter social media posts.

\section{Conclusion}
\label{section:conclusion}
We presented RuATD shared task, the first shared task on artificial text detection for the Russian language. As a result of the competition, 38 solutions have been obtained, solving the problem in two tasks modeled after the traditional concepts of the Turing test and authorship attribution for NLG methods.

The best solution of the shared task has achieved 
\begin{itemize}
    \item 83.0\% accuracy in a binary task setup;
    \item 65.0\% accuracy in a multi-class task setup.
\end{itemize}
The shared task dataset, codebase, human evaluation results, participant solutions, and other materials are now available online under Apache 2.0 license\footnote{\href{https://github.com/dialogue-evaluation/RuATD}{https://github.com/dialogue-evaluation/RuATD}}. 

The competition problem can be further treated as a Turing test in a non-interactive setting.
First of all, its direct methodological extensions are possible in such applied areas as: 
\begin{itemize}
    \item dialogue systems, and
    \item applications for editors and writers.
\end{itemize}

Another direction for future work is to conduct a critical study on the human evaluation guidelines on artificial text detection, which is still an open methodological question in the field~\cite{karpinska-etal-2021-perils}. We welcome the communities of NLP developers, linguists, and engineers to contribute to further research in the area and next criteria formulations.

\section*{Acknowledgments}
The experiments were partially carried out on computational resources of HPC facilities at HSE University \cite{kostenetskiy2021hpc}. Ekaterina Artemova and Marat Saidov were supported by the framework of the HSE University Basic Research Program.

\bibliography{dialogue.bib,anthology.bib}
\bibliographystyle{dialogue}
\clearpage

\appendix
\setcounter{figure}{0}

\begin{figure*} 
\section{Annotation Protocols} \label{section:appendix}
    \includegraphics[width=0.9\linewidth]{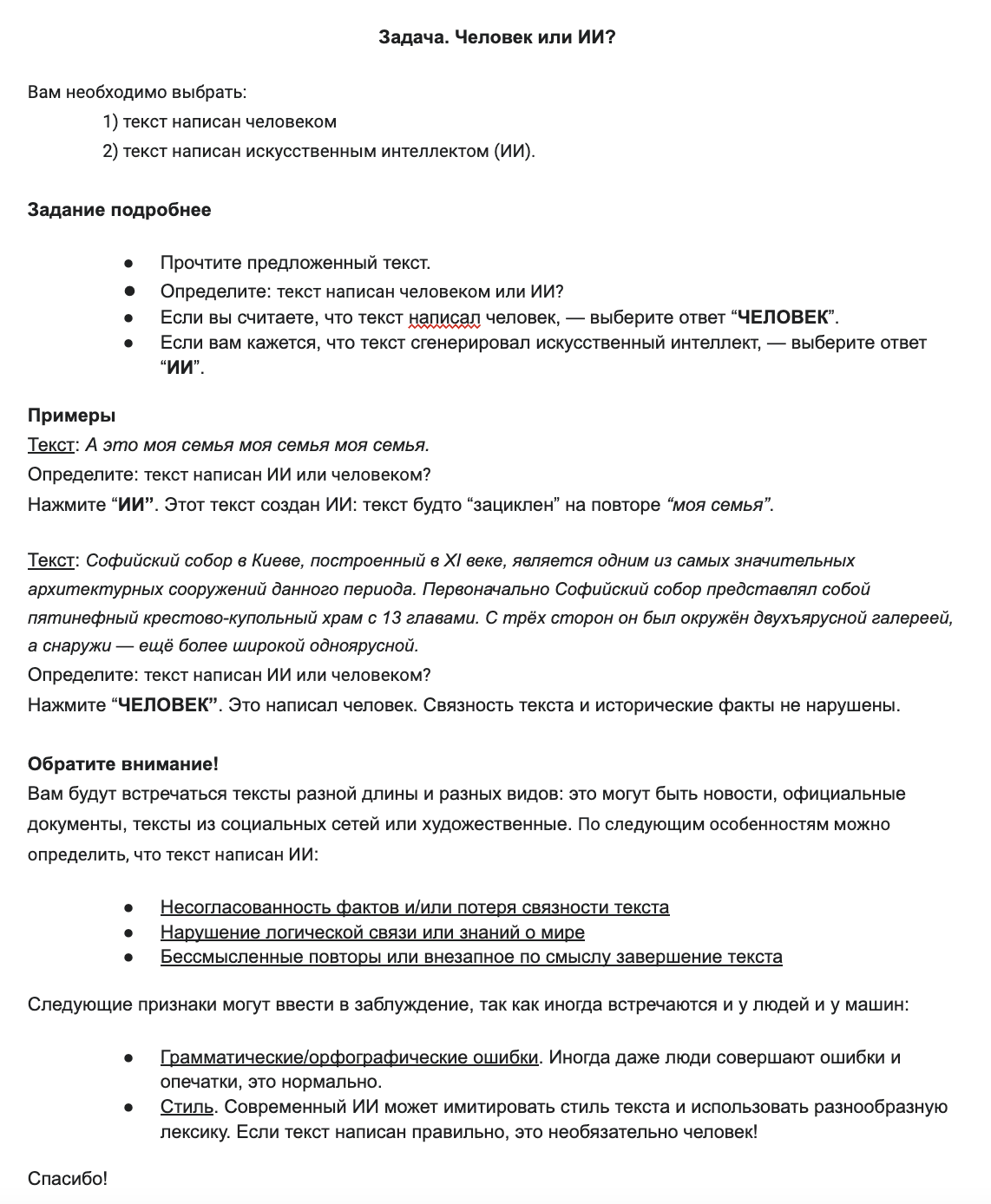}
    \caption{An example of the annotation instruction for the human evaluation.}
    \label{fig:toloka_instruction}
\end{figure*}

\end{document}